\documentclass[letterpaper, 10 pt, conference]{ieeeconf}  

\IEEEoverridecommandlockouts                              
                                                          
\usepackage{hyperref}
\usepackage{graphicx} 
\graphicspath{{./figures/}}
\usepackage{amsmath} 
\usepackage{amssymb}  
\usepackage{bbm}
\usepackage{verbatim} 
\usepackage{algpseudocode}
\usepackage{tabulary}
\usepackage{algorithm} 
\usepackage[utf8]{inputenc} 
\usepackage{subcaption} 
\usepackage{wrapfig}
\usepackage{eso-pic}

\usepackage{color}

\title{\LARGE \bf Adaptive Prior Selection for Repertoire-based Online Adaptation in Robotics}

\author{Rituraj Kaushik, Pierre Desreumaux, Jean-Baptiste Mouret$^*$
\thanks{*Corresponding author: {\tt\small jean-baptiste.mouret@inria.fr}}
\thanks{All authors have the following affiliations:}
\thanks{-Inria, CNRS, Universit\'e de Lorraine, LORIA, F-54000 Nancy, France}
\thanks{This work received funding from the European Research Council (ERC) under the European Union’s Horizon 2020 research and innovation programme (GA no. 637972, project ``ResiBots''), the Chist-Era project ``HEAP'' and the Lifelong Learning Machines program (L2M) from DARPA/MTO under Contract No. FA8750-18-C-0103.}
\thanks{Video: \url{http://tiny.cc/aprol_video}}
}

\newcommand{\algo}{APROL} 

\begin{document}
\AddToShipoutPicture*{\put(0,740){\parbox[b][\paperheight]{\paperwidth}{%
\vfill
\centering\footnotesize
R. Kaushik, P. Desreumaux, and J.-B. Mouret, ``Adaptive prior selection for repertoire-based online adaptation in robotics,'' \\
\textit{Frontiers in Robotics and AI,} vol. 6, p. 151, 2020. 
}}}
\maketitle
\IEEEpeerreviewmaketitle

\begin{abstract}
Repertoire-based learning is a data-efficient adaptation approach based on a two-step process in which (1) a large and diverse set of policies is learned in simulation, and (2) a planning or learning algorithm chooses the most appropriate policies according to the current situation (e.g., a damaged robot, a new object, etc.). In this paper, we relax the assumption of previous works that a single repertoire is enough for adaptation. Instead, we generate repertoires for many different situations (e.g., with a missing leg, on different floors, etc.) and let our algorithm selects the most useful prior. Our main contribution is an algorithm, \algo{} (Adaptive Prior selection for Repertoire-based Online Learning) to plan the next action by incorporating these priors when the robot has no information about the current situation. We evaluate \algo{} on two simulated tasks: (1) pushing unknown objects of various shapes and sizes with a robotic arm and (2) a goal reaching task with a damaged hexapod robot. We compare with ``Reset-free Trial and Error'' (RTE) and various single repertoire-based baselines. The results show that \algo{} solves both the tasks in less interaction time than the baselines. Additionally, we demonstrate \algo{} on a real, damaged hexapod that quickly learns to pick compensatory policies to reach a goal by avoiding obstacles in the path. 
\end{abstract}


\section{Introduction} 
Reinforcement Learning (RL) algorithms have achieved impressive successes during the last few years, from learning to play games from pixels to beating professional Go players, but at the expense of enormous interaction time with the system. For example, they required up to 38 days of game-play (real-time) for Atari 2600 games~\cite{mnih_human-level_2015}, 4.8 million games for Go~\cite{silver_mastering_2016}, or about 100 hours of simulation time (more if real-time) to train a 9-DOF mannequin to walk~\cite{heess2017emergence}. This makes these algorithms suitable only for policy \emph{synthesis}, that is, creating a policy for a robot in simulation, but impossible to use for \emph{online learning} in robotics, that is, adapting online to a new system or a new situation. By the term ``situation'' we refer to any perturbation in the dynamics of the robot caused by physical damages, faults in the actuators or environmental changes such as terrain condition. 


\begin{figure*}[h!]
  \centering
  \includegraphics[width=0.8\linewidth]{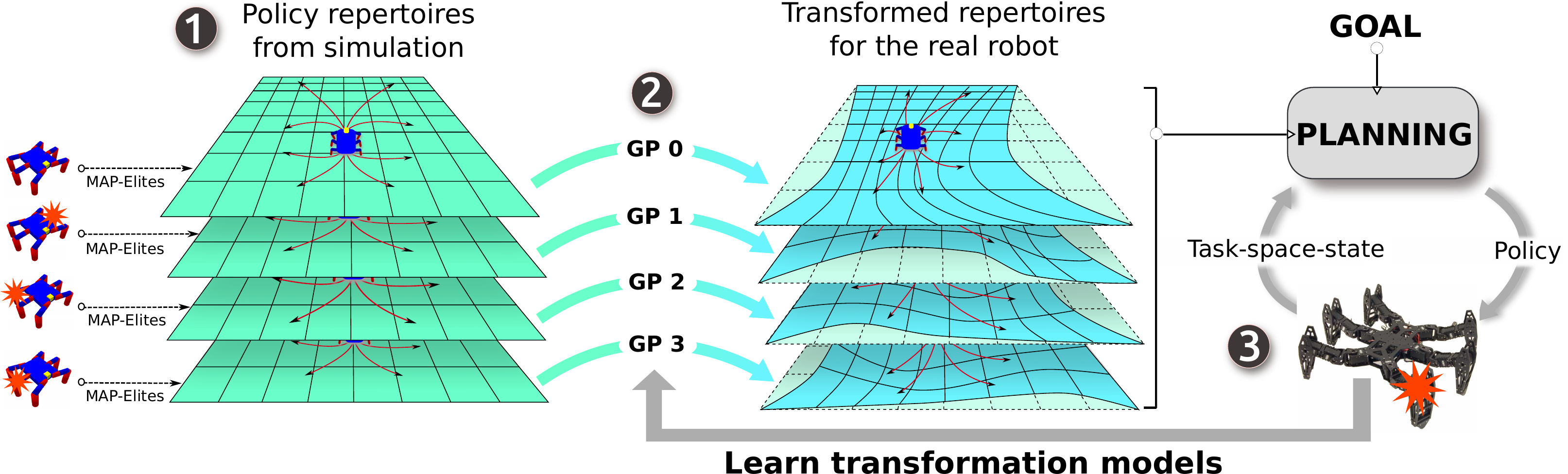}
  \caption{\small \label{fig_1_concept_digram_v3} Overview: \algo{} uses multiple repertoire-based priors for fast online adaptation in unforeseen situations. \textbf{(1)} First, in a low fidelity simulator, we generate multiple repertoires of elementary policies for various situations of the robot, such as, being damaged, slippery floor, interaction with a novel object etc. A repertoire is basically a discrete one-to-one association between elementary policies and their corresponding task-space transitions on the robot. \textbf{(2)} Then, we use those repertoires as prior mean-function for Gaussian process regression models that learn to transform the task-space transitions in each of the repertoires to the task-space transitions of the real robot using the past observations. \textbf{(3)} Finally, given the goal, \algo{} iteratively picks the most suitable elementary policies from the most suitable repertoire at every replanning step to reach the goal in a minimum number of steps. In between every replanning step, the task-space transformation models are updated with the past observations.} 
\end{figure*}  

Model-based reinforcement learning algorithms (MBRL) allow robots to learn policies with less interaction time by alternating between learning a dynamical model of the robot from the observed data, and using that model either for finding a policy~\cite{deisenroth2011pilco, chatzilygeroudis2017black,kaushik2018multi} or for model predictive control~\cite{williams2017information,nagabandi2018learning,chua2018deep}. Since these algorithms optimize a policy (or plan an action) using the learned dynamical model, they can be highly data-efficient. However, MBRL does not scale well with the dimensionality of the state-space as the amount of data required to learn a model typically scales exponentially with the dimensionality of the input space \cite{keogh2010curse}.

A promising way to address the ``curse of dimensionality'' in reinforcement learning \emph{for online adaptation} is to learn a model in ``task-space'' of the robot that predicts how the outcomes of elementary policies stored in a \emph{repertoire} change in reality compared to the simulated robot \cite{cully_evolving_2015,cully_robots_2015,chatzilygeroudis2018reset,duarte2017evolution,sharma2019dynamics}. By the term ``task-space'' we refer to the space where the operation of the robot is required (e.g., it can be the x and y coordinate positions for a mobile robot). This process splits the policy search problem in two parts: first, search for a repertoire of policies in simulation, where large number of interactions are possible; and second, on the real robot, search for the most appropriate policy from the repertoire, which is usually easier than directly searching on the original policy parameter space. For instance, given a repertoire of elementary policies that makes a 6-legged robot (18D joint space) walk in different directions (i.e, one policy for each walking direction on the 2D plane/task-space), we can learn a model to predict how this 2D task-space is transformed (i.e, a mapping from expected transitions to the observed transitions of the robot on the surface) when these policies are transferred to a damaged robot (e.g, one missing leg) \cite{chatzilygeroudis2018reset} (Figure~\ref{fig_2_repertoire_based_learning}). With an accurate prediction model, a planning algorithm can then select a sequence of these elementary policies from the repertoire by taking into account the outcome difference between the prior --- the intact robot in simulation --- and the reality --- the damaged robot. For instance, for a mobile robot, this approach might learn a transformation model in the 2D center of mass (COM) position-space. In that case the input to the model is the change in COM positions expected by the repertoire for the associated policy and the target is the corresponding real observation on the robot. Since the dimension of the task-space is often much smaller than the state-space, this reduces the dimensionality of the model, and therefore the amount of interaction time.

Like with any learning algorithm based on prior knowledge, the effectiveness of the adaptation process depends critically on the difference between the prior and the reality: the bigger the difference, the worse it will perform (more interaction time, lower quality policies). In this paper, we address this issue by allowing the adaptation algorithm to select the most interesting prior among a set of priors learned beforehand. In other words, we learn several repertoires of elementary policies in simulation, each with a unique situation (e.g., different damages), and the robot adapts by both searching for the most suitable prior (i.e the repertoire) and correcting their expected outcomes using a model learned from the observations.

To do so, we propose to evolve several repertoires of elementary policies for the robot using an evolutionary algorithm called MAP-Elites~\cite{cully_robots_2015,mouret_illuminating_2015} (in simulation), each with a unique situation picked from a sub-set of probable situations that the robot might face in reality. Each of these repertoires associates different task-space transitions (e.g relative displacement) with unique elementary policies. Using each of these repertoires as ``prior mean-functions'' for Gaussian processes (GP) regression models~\cite{rasmussen2006gaussian}, we learn as many models as the number of repertoires from the observations on the real robot. Each of these models maps expected task-space transitions stored in the corresponding repertoire to the actual task-space transition on the robot. More concretely, we iteratively learn a probabilistic model that predicts how these repertoires transform themselves when applied on the real robot as we collect more data from the real robot. Then, instead of selecting a single global model for controlling the robot, we pose this as a maximum a posteriori (MAP) estimation problem of selecting the next elementary policy from one of the repertoires, given the repertoires, the past observations and the goal. We call this algorithm \algo{} (Adaptive Prior selection for Repertoire-based Online Learning) (Figure \ref{fig_1_concept_digram_v3}). The main novelty in \algo{} compared to the previous work is that it uses multiple repertoires instead of one and adapt online by automatically selecting the most suitable policy from one of those repertoires based on the current situation. 

\section{Related Work}
\subsection{Data-Efficient Learning in Robotics}
To be useful for online learning in robotics, the algorithm should allow the robot to learn within a very short interaction time (ideally less a few minutes) \cite{chatzilygeroudis2018survey}. In this direction, MBRL (Model-Based Reinforcement Learning) algorithms showed promising results by allowing simple robots to learn new skills within a few minutes of interaction with the real world \cite{kaushik2018multi,chatzilygeroudis2017black,deisenroth_gaussian_2015,chua2018deep}. These model-based approaches mainly fall into two categories depending upon where the learning process is inserted \cite{chatzilygeroudis2018survey}: (1) alternating between learning a model of the dynamics and learning an optimal policy according to the model, which is called model-based policy search \cite{deisenroth_gaussian_2015,chatzilygeroudis2017black}, and, (2) learning a model of the system dynamics, then using it along with a planner or a Model-Predictive Control (MPC) loop  \cite{nagabandi2018learning,chua2018deep,williams2017information}, which is often called adaptive model predictive control. Most of the experiments so far have been based on episodic learning: after each trial, the robot is reset to the same starting state. While this makes sense for manipulation tasks, which can be reset easily, it is difficult to use for locomotion tasks in the field.

While MBRL algorithms are more data-efficient than model-free policy search approaches, learning a model that is good enough to plan and control a complex robot requires a large amount of data/observations. This contrasts with animal behavior which can adapt to new situations (such as uneven terrain, broken limbs) within a minute or even in seconds. To accelerate the learning process and thereby increase the data-efficiency, many recent papers propose to leverage prior knowledge about the system dynamics; such as, using a known but low fidelity simulator or a parametric mathematical model of the dynamics. In traditional robotics, data efficiency is achieved by simply identifying the parameters of such mathematical models using the observed data from the real robot \cite{Hollerbach2016}. In more recent approaches \cite{chatzilygeroudis2018using,cutler_efficient_2015}, a parametric \cite{chatzilygeroudis2018using} or fixed \cite{cutler_efficient_2015} model is ``corrected'' with a non-parametric model to capture potentially non-linear effects. 

Recently, meta-learning models of the dynamics showed promising results for fast adaptation to new situation \cite{nagabandi2018learning}. Typically, they optimize the initial parameters for a neural network based model of the dynamics of the robot such that the model can be adapted quickly to match the true model of the robot with a small number of observations. The main challenges of this kind of meta-learning approaches is that (1) it is much more computationally demanding than simply learning the model (since each model needs to be evaluated on its capacity to learn instead of its performance, and (2) they assume that a single, well chosen parametrization of the model will be a good starting point for any future change, which is not guaranteed to be true.
 
\begin{figure}[h!]
  \centering
  \includegraphics[width=1.0\linewidth]{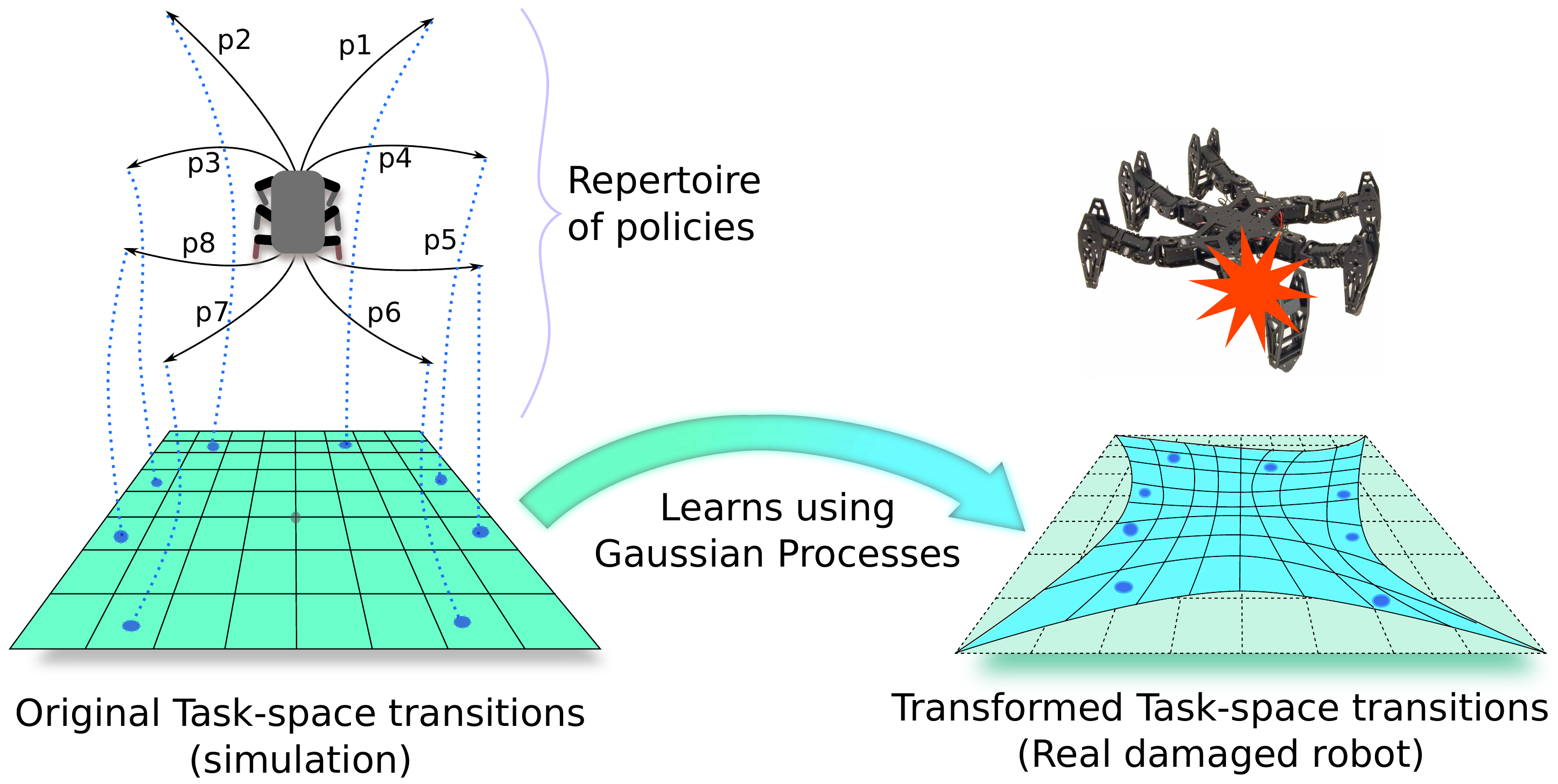}
  \caption{\small \label{fig_2_repertoire_based_learning} Repertoire-based learning in robotics: First, a repertoire of elementary policies is evolved for the robot using a known but imperfect simulator. This repertoire associates potentially every discretized task-space (or outcome-space) transitions to unique elementary policies. Then, during deployment of the real robot, a Gaussian process model is learned which transforms this ``prior'' task-space in such a way that the outcomes of policies on the real robot match with the new transformed task-space. RTE~\cite{chatzilygeroudis2018reset} uses this model with Monte-Carlo tree search to pick the policies in a sequential manner from the repertoire to solve the target task.}
\end{figure}

 
\subsection{Gaussian Process Regression with Non-Constant Prior}
\label{sec:gp}
\begin{figure*}[h!]
  \centering
  \includegraphics[width=0.9\textwidth]{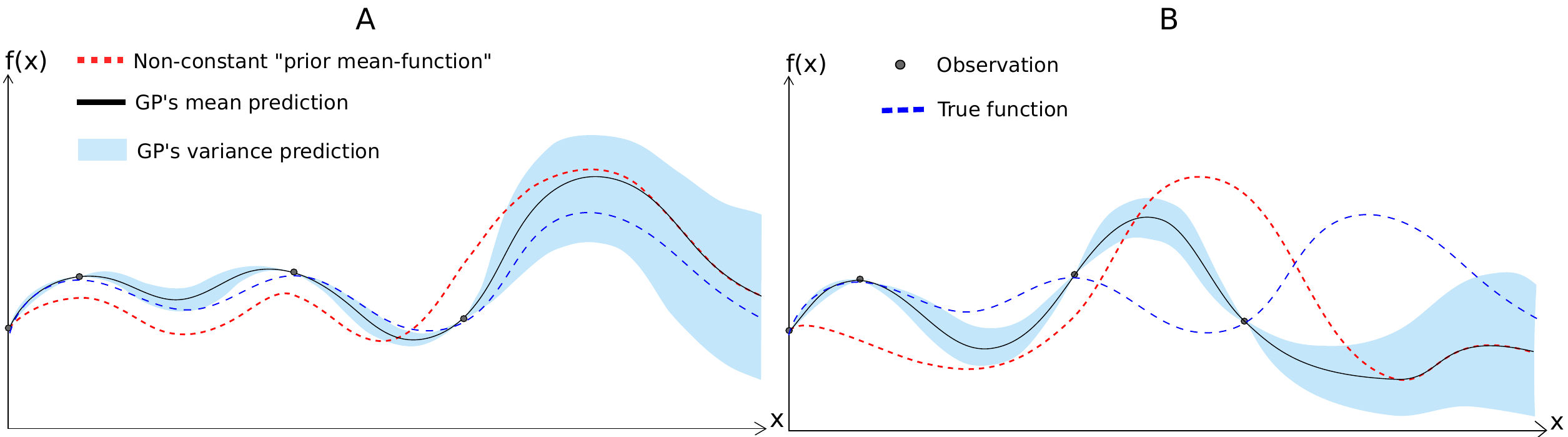}
\caption{\small \label{fig_3_gp_prior} Gaussian processes regression with non-constant prior - Plots show how the model fitting is impacted by the selection of prior mean-function. On the left (plot A), the prior mean function is more similar to the true underlying function that the model tries to fit with 4 given data points. On the right (plot B), the prior mean-function is very different from the true underlying function. As as result, in plot A, the model fitting is very close to the true function. However, for plot B, due to selection of ``wrong prior'', model fitting is far from the true function.}  
\end{figure*}

One of the key elements of data-efficiency in MBRL is the ability to use a prior \cite{chatzilygeroudis2018survey} that comes from simulation \cite{cully_robots_2015,cutler_efficient_2015, papaspyros2016safety, pautrat2018bayesian,chatzilygeroudis2018reset,chatzilygeroudis2018using,nagabandi2018learning,Saveriano2017DataefficientCP}, either directly or via meta-learning. Most previous algorithms leverage Gaussian processes (GP) \cite{rasmussen2006gaussian,chatzilygeroudis2018survey,deisenroth_survey_2013} as data-driven models because they work well with a few data and because it is easy to introduce priors (easier than in neural networks).

A GP is an extension of multivariate Gaussian distribution to an infinite-dimension stochastic process for which any finite dimensions will be a Gaussian distribution~\cite{rasmussen2006gaussian}. It is a distribution over functions, specified by mean function $\mu(\cdot)$ and covariance function $k(\cdot,\cdot)$:
\begin{align}
  f(\mathbf{x}) \sim GP(\mu(\mathbf{x}), k(\mathbf{x}, \mathbf{x}^\prime))
\end{align}

As mentioned before, compared to neural networks, a GP can easily include prior knowledge about the underlying function~\cite{rasmussen2006gaussian}. In particular, we can provide a prior ``mean-function'' to the GP model which is our prior belief about the prediction when no data is available to train the model (Figure~\ref{fig_3_gp_prior}). If $\mu(\mathbf{x})$ is the prediction mean and $\sigma^2(\mathbf{x})$ is the prediction variance of a GP model for any input $\mathbf{x}$, $M(\mathbf{x})$ is the prior belief about the prediction mean of the model for the same input $\mathbf{x}$, $\sigma_{n}^2$ is the prior noise and $D_{1:t}$ is the set of $t$ observations, then the GP is computed as follows:
\begin{align}
  & P(f(\mathbf{x}) | D_{1:t}) = \mathcal{N}(\mu(\mathbf{x}), \sigma^2(\mathbf{x})) \\ & \text{where,} \\
  & \mu(\mathbf{x}) = M(\mathbf{x}) + \boldsymbol{k}^T(\mathbf{K} + \sigma_n^2 I)^{-1} (D_{1:t} - M(\mathbf{x}_{1:t})) \\
  & \sigma^2(\mathbf{x}) = k(\mathbf{x}, \mathbf{x}) - \boldsymbol{k}^T(\mathbf{K} + \sigma_n^2 I)\boldsymbol{k}
\end{align}

Where, $\mathbf{K}$ is the kernel matrix with entries $\mathbf{K}[i,j] = k(\mathbf{x}_i, \mathbf{x}_j)$ and $\boldsymbol{k} = k(D_{1:t}, \mathbf{x})$. As mentioned above, here $k(\cdot, \cdot)$ is the covariance function or the kernel function.

In Bayesian optimization applied to robot learning, there are many recent work that use non-constant priors coming from simulation to model the cost function using GP \cite{cully_robots_2015, papaspyros2016safety, pautrat2018bayesian}. In particular, \cite{pautrat2018bayesian} uses several repertoires of policies evolved for different situations that perform the same task, but in different ways (i.e., different behaviors). Then it uses the performance scores stored in each of the repertoires as prior mean-functions to the GP to learn the performance function for Bayesian optimization on the real robot. In effect, it learns as many models as the number of repertoires. To select a policy, a novel acquisition function called MLEI (Most Likely Expected Improvement) is used which considers the likelihood of the prior being close to the reality while computing expected improvement of performance for a policy.  

Many recent work also used priors from a simulator to learn a ``residual model'' with GP, i.e, the difference between the simulated and real robot instead of learning the system model from scratch. For example, model-based policy search algorithm like PILCO \cite{deisenroth2011pilco} or Black-DROPS \cite{chatzilygeroudis2017black} can be combined with simulated priors and learn to control a cart-pole in 2 to 5 trials \cite{cutler_efficient_2015, Saveriano2017DataefficientCP,chatzilygeroudis2018using}.

On the one hand, these contributions prove that using well-chosen priors with GPs is a promising approach for data-efficient learning; on the other hand all previous algorithms assume that we know a good prior in advance. This is a very strong and crucial assumption as a misleading prior can substantially increase interaction time needed to learn to control the system. In this paper, we relax this assumption and argue that by using multiple priors for a subset of the possible set of situations and allowing the algorithm to choose the best one for modeling and planning improves online learning and adaption for robotics. 

\subsection{Repertoire-based Learning in Robotics}

Repertoire-based approaches also use prior knowledge from simulation to make learning more data-efficient. Their key principle is to learn a large and diverse set of policies in simulation with a ``quality diversity'' algorithm \cite{mouret_illuminating_2015,pugh2016quality,cully2018quality}, then use an optimization or search process to pick the policies that works best in current situation \cite{cully_evolving_2015,cully_robots_2015,chatzilygeroudis2018reset,duarte2017evolution,sharma2019dynamics}. The most prominent algorithm of this family is ``Intelligent Trial-and-error (IT\&E)''~\cite{cully_robots_2015}. Before deployment a repertoire of policies for an intact robot is evolved in simulation using the MAP-Elites~\cite{mouret_illuminating_2015} algorithm such that many alternative but good ways of performing the task are found. For instance, there are many ways of walking with a 6-legged robot: a tripod gait with the 6 legs, a jumping gait in which all the gaits are used simultaneously, a limping gait with only 5 legs, etc. For each of these families  of gaits, there exist a well-optimized gait that performs the tasks (walking) in a specific way. If an adaptation is needed, IT\&E searches for the most appropriate gait in the repertoire using Bayesian optimization, that is, it models the performance function on the real robot with GP and uses the uncertainty of the GP prediction to balance exploration and exploitation during the search process.  Importantly, IT\&E uses the performance scores computed in simulation as the ``prior mean-function'' for the GP model (section sec. \ref{sec:gp}) so that the robot has initial ``guesses'' about the performance of each policy of the repertoire. Thanks to this 2-step learning process, IT\&E allows a damaged 6-legged robot (12D joint space) to find out compensatory policies (36D space) within 2 minutes of interaction with the robot. However, since the robot has to be reset back to the original state after each episode (this is an episodic learning algorithm), IT\&E cannot be used ``as is'' to adapt on the field. 

The repertoire-based learning algorithm that is the closest to the present work is ``Reset-free Trial and Error'' (RTE) \cite{chatzilygeroudis2018reset}. Like in IT\&E, RTE searches for a repertoire of diverse policies using MAP-Elites. However, instead of searching for many ways of performing a single task, the repertoire captures the best way of performing many variants of the task. In the hexapod robot case, each policy reaches a different points around the current positions of the robot; for instance, one policy to walk forward, one policy to walk backward, one policy to turn right by 30 degrees, etc. On a damaged robot, this repertoire needs to be modified since policies that are supposed to make the robot move forward do not lead to the same behavior anymore due to the damage as well as the reality gap with the simulator. For instance, a policy that is supposed to make the robot move forward might make it turn right because of a damage to a leg.

To adapt this repertoire during deployment, RTE learns a Gaussian process model to predict how the expected outcomes of these policies change on the real robot. For instance, it learns that the policy that was supposed to move forward actually makes the robot turn right, which is useful if the robot needs to turn right. More precisely, this model is a probabilistic transformation from the expected outcomes to observed outcomes using the data obtained during execution of the elementary policies on the robot (Figure \ref{fig_2_repertoire_based_learning}). To choose the next policy, RTE uses Monte Carlo Tree Search (MCTS), that is, a planning algorithm, so that it can exploit the uncertainty predictions of the GP when planning a sequence of elementary policy (e.g., move forward for 30 cm, turn right for 30 degrees, etc.). Note that, similar to IT\&E, RTE also uses the outcomes of the simulator (stored in the repertoire) as prior mean-function for the GP model, which makes the learning process very data-efficient. RTE showed that a damaged hexapod (single blocked leg) robot can recover $77.52\%$ of its capability through online adaptation compared to an intact robot. 

Using a repertoire of pre-learned policies assumes that the repertoire contains some policies that will work in the current situation (e.g., with a damaged robot). However, since the repertoires are evolved without anticipating different situations the robot might face in reality, it is possible that none of the policies work (although experiments show that a single repertoire allow robots to adapt to surprisingly many situations \cite{cully_robots_2015}). A way to relax this assumption is to provide many repertoires (several dozens) and make the algorithm choose the most likely one according to the observations. This approach was recently proposed for Bayesian optimization, that is, for episodic learning, by \cite{pautrat2018bayesian}. To do so, \cite{pautrat2018bayesian} introduced a new acquisition function that combines the likelihood of the repertoire given the observations (how well the prior matches the observations) and the expected improvement (how much we would gain if the policy $\pi$ works as expected on the robot). Thanks to this algorithm, \cite{pautrat2018bayesian} showed that a damaged hexapod (in simulation) can learn to climb stairs with less than 10 trials by using multiple repertoires generated for various damage conditions and various stair heights. The present work follows a similar line of thought but extends it to the reset-free learning approach introduced by the RTE algorithm.

\section{Problem Formulation}

We consider a system whose transition in the task-space depends not only on the current task-space state and the policy, but also on the current situation (e.g, icy vs rocky terrain for mobile robot). Then, task-space transition dynamics of such systems can be written as: 
\begin{align}
s_{t+1} = s_{t} + F (s_{t}, \pi_\theta, c) + w \label{eq:task_space_dynamics}
\end{align}

where, $s_t$, $s_{t+1}$ are the locations in task-space at time-step $t$ and $t+1$ respectively,
$\pi_\theta$ is the open-loop policy/controller parameterized by $\theta$, $c \in \mathbb{C}$ specifies the current situation, $w$ is the Gaussian system noise and $F(\cdot,\cdot,\cdot)$ is the task-space transition function of the system. Note that, $\mathbb{C}$ can potentially be an infinite set, which means that the system can face infinitely possible situations during its deployment that can change its transition dynamics.

We assume that we neither have access to $F(\cdot,\cdot,\cdot)$ nor have knowledge about the current situation $c$. However, we have access to a low fidelity simulator of the system $\hat{F}(\cdot,\cdot,\cdot)$ and a set $\mathbb{C'} \subset \mathbb{C}$ of probable situations that the system might face during deployment. The goal is to drive the system from a starting task-space state $s_0$ to the target task-space state $s_g$ in a minimum number of steps by executing a sequence of elementary policies. Here, elementary policies are open-loop policies that are applied for a short period of time (a few seconds) on the robot which cause a small change in the task-space state of the robot.

In other words, we consider a robot that follows eq. \ref{eq:task_space_dynamics} and might face any situation, such as broken joints, slippery floor or a novel object to manipulate during its mission. These situations cannot be predicted beforehand and the robot is not equipped with any specialized sensor either to observe such situations. Now, if such situations arise, instead of aborting its mission, the robot has to figure out a sequence of compensatory policies to continue its mission and accomplish the goal as quickly as possible.

\section{Approach}
\subsection{Overview}
\algo{} allows a robot to ``learn while doing'' instead of ``learning and then doing''. Our approach is based on 3 main stages (Figure \ref{fig_1_concept_digram_v3}): 

\begin{enumerate}
  \item Before deployment of the robot, several repertoires of elementary policies are generated for the robot with an evolutionary algorithm called ``MAP-Elites'' using a relatively low-fidelity simulator of the real robot (section \ref{sec:repertoires}). Each of these repertoires are generated for a unique situation or circumstance that the robot might face during its mission, such as a broken limb, a novel object to interact with or different terrain conditions etc. Each of these repertoires is basically a one to one association between the evolved elementary policies and the corresponding transitions they cause on the robot.  
  
  \item At every replanning-step, using the past observations from the real robot, we learn a probabilistic mapping $g:S \mapsto S$, where $S$ is the task-space, for each of the repertoires to predict how the task-space-transitions in the repertoires transform themselves when corresponding policies are applied on the real robot (section \ref{sec:learning_gp}). This transformation models are learned using GP, because (i) we can set a prior mean-function for the GP which is the prior belief about the underlying function that the GP needs to fit, (ii) instead of deterministic prediction, GP outputs a probability distribution, which allows us to incorporate model uncertainty in the planning stage. For every repertoire, we learn this model using the expected outcomes of the repertoire itself as prior mean-function.
  
  \item Using the past observations and the given goal, \algo{} picks the best policy from one of the repertoires and applies it on the robot for one replanning-step (section \ref{sec:planning}). Then the process repeats from step 2.
\end{enumerate}


\begin{algorithm}[H]
  \footnotesize
  \caption{Generate Priors}
  \label{algo:priors}
  \begin{algorithmic}[1]
    \Require $S\in\mathbb{R}^{n_s}$ \Comment{Task-space}
    \Require $\Theta \in \mathbb{R}^{n_\theta}$ \Comment{Policy space}
    \Require $C = \{c_0, c_1,...,c_n\}$ \Comment{Probable situations}
    \Require $bot\_sim$ \Comment{Simulator of the robot}
    \Require $f_{eval}(.)$ \Comment{Performance function}
    \Require $N_{max}$ \Comment{Max evaluation}
    \State $\mathcal{R} = \{\}$ \Comment{Empty set of Repertoires}
    \For {$c$ in C}
      \State $\Pi = map\_elites(bot\_sim,\Theta,c,S,f_{eval},N_{max})$
      \State Insert repertoire $\Pi$ in $\mathcal{R}$  
    \EndFor
    \State Return $\mathcal{R}$
  \end{algorithmic}
\end{algorithm}

\subsection{Generating Repertoire-Based Priors}
\label{sec:repertoires}
We assume that the robot can be controlled by a low-level elementary policy $\pi_{\boldsymbol\theta}$ (typically, an open-loop policy) parameterized by $\boldsymbol\theta \in \mathbb{R}^{n_\theta}$ and that any point on the task-space can be described by a vector $\mathbf{s} \in \mathbb{R}^{n_s}$.
In simulation, the task-space transition caused by the policy $\pi_{\boldsymbol\theta}$ can simply be written as $\Delta \mathbf{s}_\theta$. Additionally, we assume that a set $\mathbb{C'}=\{c_0, c_1,..,c_n\}$, which is a subset of all the possible situations $\mathbb{C}$ is available. Then, for each situation $c \in \mathbb{C'}$, we use an iterative algorithm called ``MAP-Elites'' ~\cite{cully_robots_2015,mouret_illuminating_2015, vassiliades2017using} to evolve a repertoire of elementary policies in simulation, such that a wide range of task-space transitions can be captured in the repertoire~\cite{cully_robots_2015,cully_evolving_2015, Duarte2018EvolutionOR} (see Algorithm~\ref{algo:priors}). Nevertheless, other quality diversity algorithms \cite{pugh2016quality,cully2018quality} could also be used to generate the repertoire with almost no influence on the behavior of \algo{}.

To start with, MAP-Elites discretizes the task-space into some regions or cells, each of which is identified using a cell identifier ($cell\_id$), which is a unique key to specify a cell. At the beginning, MAP-Elites randomly initializes some policies and test them in simulation to find out the task-space transitions $\Delta \mathbf{s}_\theta$ and their performance score $r_{\boldsymbol\theta}$. Then, they are included in the repertoire as tuples of policies, their corresponding transition, the performance score and the cell id as $(\pi_{\boldsymbol\theta}, \Delta \mathbf{s}_\theta, r_{\boldsymbol\theta}, cell\_id)$. Here, the performance is a user defined function with which some constraints can be imposed on the behavior of the robot. For example, we can set a lower performance score for a policy if it produces higher joint torques on the joints. That way MAP-Elites will prefer the policy with lower torque if two policies produce same task-space transition on the robot. After this initialization, MAP-Elites performs the following three steps iteratively until the maximum number of valuations are reached:

\begin{enumerate}
  \item Randomly picks a tuple from the repertoire and adds a small random variation to the policy.
  \item Simulates the policy to get the task-space transition, performance score and the $cell\_id$ to create a new tuple.
  \item Inserts the new tuple into the repertoire if no tuple exists with the same $cell\_id$, or, replaces an existing tuple with the same $cell\_id$ but with a lower performance score (discards the new tuple otherwise).  
\end{enumerate}

Thus, each repertoire is a set of tuples $(\pi_{\boldsymbol\theta}, \Delta \mathbf{s}_\theta, r_{\boldsymbol\theta}, cell\_id)$, where, no two tuples have the same $cell\_id$. One thing to be noted here is that although MAP-Elites is computationally expensive, it can be parallelized on large clusters to compute the repertoires before deployment of the robot. It is worth mentioning here that we use CVT variant of MAP-Elites~\cite{vassiliades2017using} that uses centroidal voronoi tesselation to discretize the task-space into the user specified number of homogeneous geometric regions. In CVT Map-Elites, the number of cells remains fixed irrespective of the dimensionality of the task-space, making it scalable to a very high dimensional task-space.

\begin{algorithm}[H]
  \footnotesize
  \caption{Planning using \algo}
  \label{algo:planning}
  \begin{algorithmic}[1]
    \Require $\mathcal{R}=\{\Pi_0\, \Pi_1,...,\Pi_n\}$ \Comment{Set of repertoires}
    \Require $G$ \Comment{The task/goal}
    \Require $task\_planner$
    \Require $D=\phi$  \Comment{Empty observations set}
    \State $T=\{gp_0, gp_1,\cdots, gp_n \}$ \Comment{Initialize models}
    \While {task not solved}
      \State $s = get\_current\_state()$
      \State $s_g = task\_planner(G,s)$ \Comment{Current sub-goal}
      \State $\pi^* = \operatorname*{argmax}_{\pi \in \Pi, \Pi \in \mathcal{R}} P( s_g | \pi, D, \Pi) P(\Pi|D) P(\pi)$
      \State Execute $\pi^*$ and record data in $D$
      \State Update transformation models $T$ with $D$
    \EndWhile
  \end{algorithmic}
\end{algorithm}

\subsection{Learning the Transformation Models with Repertoires as Priors}
\label{sec:learning_gp}
Here, we use the same approach that RTE \cite{chatzilygeroudis2018reset} used to learn the transformation model with the repertoire as priors (see Figure \ref{fig_2_repertoire_based_learning}). Since the policies in the repertoires come from a simulator, how they change the state of the system is an approximation of the reality. Moreover, if the real situation of the robot (e.g, different floor conditions, novel object to interact with, mechanical damage etc.) is different from those of the repertoires, then the corresponding transitions for the policies will not align perfectly with the real system. However, transitions that we observed in the simulation can be a ``prior'' (i.e., prior in Bayesian model learning) to learn the actual transitions we see on the real system, provided the situation of the robot in simulation is somewhat close to the real situation. Thus, we use GPs to learn the transformation models of the task-space from simulation to reality, where we use the transitions stored in the repertoires themselves as prior mean-functions to these models.

Suppose, an arbitrary policy $\pi_{\boldsymbol\theta}$ from a repertoire produces a task-space transition $\Delta \mathbf{s}_\theta$ on the simulated robot. Now suppose, the same policy $\pi_{\boldsymbol\theta}$ produces transition of $\Delta \mathbf{s}_{\theta, real}$ on the real robot. We can learn a model to predict this transformation of $\Delta \mathbf{s}_\theta$ to $\Delta \mathbf{s}_{\theta, real}$. More concretely, we learn a probabilistic model using GP, where input is $\Delta \mathbf{s}_\theta$, the target is $\Delta \mathbf{s}_{\theta, real}$ and the corresponding prior mean is $\Delta \mathbf{s}_\theta$ itself. If $f(\Delta \mathbf{s}_\theta)$ is the function that transforms the task-space from simulation to reality, then for each prediction dimension $d=1,2,...n$, the GP can be computed as:
\begin{align}
  & P(f_d(\Delta \mathbf{s}_\theta) | D_{1:t}^d) = \mathcal{N}(\mu_d(\Delta \mathbf{s}_\theta), \sigma_d^2(\Delta \mathbf{s}_\theta))
\end{align}
where,
\begin{align}
  & \mu_d(\Delta \mathbf{s}_\theta) = M_d(\Delta \mathbf{s}_\theta) + \boldsymbol{k}_d^T(\mathbf{K}_d + \sigma_n^2 I)^{-1} \nonumber\\
  & \text{~~~~~~~~~~~~~~~}(D_{1:t}^d - M_d({\Delta \mathbf{s}_\theta}_{1:t})) \label{eq_gp_mean} \\
  & \sigma_d^2(\Delta \mathbf{s}_\theta) = k_d(\Delta \mathbf{s}_\theta, \Delta \mathbf{s}_\theta) - \boldsymbol{k}_d^T(\mathbf{K}_d + \sigma_n^2 I)\boldsymbol{k_d}
\end{align}

Where, $D_{1:t}^d = \{ f_d({\Delta \mathbf{s}_\theta}_1), f_d({\Delta \mathbf{s}_\theta}_2), ..., f_d({\Delta \mathbf{s}_\theta}_t)\}$ is the set of $d^{th}$ dimension of the observations on the real robot, $M_d(.)$ is the $d^{th}$ dimension of prior mean from the repertoires such that $M_d(\Delta \mathbf{s}_\theta) = \Delta \mathbf{s}_\theta[d]$,  $\sigma_{n}^2$ is the prior noise, $\mathbf{K}_d$ is the kernel matrix with entries $\mathbf{K}_d[i,j] = k_d({\Delta \mathbf{s}_\theta}_i, {\Delta \mathbf{s}_\theta}_j)$ and $\boldsymbol{k}_d = k_d(D^d_{1:t}, {\Delta \mathbf{s}_\theta})$. We use squared exponential kernel given by:
\begin{align}
  k_d({\Delta \mathbf{s}_\theta}, {\Delta \mathbf{s}^\prime_\theta}) = \sigma_{se}^2 \exp(-\frac{||{\Delta \mathbf{s}_\theta}-{\Delta \mathbf{s}^\prime_\theta}||^2}{l^2}) \label{eq:kernel}
\end{align}

Where, $\sigma_{se}$ and $l$ are hyperparameters. We initialize one GP model for each of the repertoires and train them iteratively as we collect more observations from the real robot during deployment.







\subsection{Model-based Planning in Presence of Multiple Priors}
\label{sec:planning}
Once GP models are initialized, the robot is deployed in the environment. We assume that the main task/goal of the robot is sub-divided into a sequence of goals in the task-space by a high-level task-planner (e.g., path planning algorithm such as A*) and the robot has to achieve the first sub-goal by applying a suitable policy from one of the repertoires. For example, for a mobile robot, the main task is to reach a particular position in the room. Then the high-level planner will sub-divide this task into a sequence of sub-goals along the shortest path, avoiding the obstacles in the room. Note here that, since high-level task-planner gives the next sub-goal for the robot to achieve, it can also be replaced with a human operator giving high-level commands (such as move left, push object right, grab etc.) remotely to the robot. At every time step, the high-level task planner re-plans the sub-goals according to the current task-space location of the system. Now, given the next sub-goal $\mathbf{s}_{g_t}$, the past observations $obs_{0:t-1}$ and the repertoires $\mathcal{R}=\{\Pi_0, \Pi_1,..., \Pi_k\}$, we can frame the next policy selection problem as a maximum a posteriori (MAP) estimation problem as follows: 

Let, $\pi_\theta$ be any elementary policy and $P(\Pi|obs_{0:t-1})$ be the probability of the repertoire $\Pi$ to match the actual situation, given the past observations. Then, the next elementary policy $\pi_{\theta_t}^*$ is given by
\begin{align}
  \pi_{\theta_t}^* &= \operatorname*{argmax}_{\pi_\theta \in \Pi, \Pi \in \mathcal{R}} P(\pi_\theta | \mathbf{s}_{g_t}, obs_{0:t-1}, \Pi) P(\Pi|obs_{0:t-1}) \nonumber \\
  &= \operatorname*{argmax}_{\pi_\theta \in \Pi, \Pi \in \mathcal{R}} \frac{P( \mathbf{s}_{g_t} | \pi_\theta, obs_{0:t-1}, \Pi) P(\pi_\theta)}{\sum_{\pi'_\theta \in \Pi', \Pi' \in \mathcal{R}} P(\mathbf{s}_{g_t}| \pi'_\theta, obs_{0:t-1}, \Pi')P(\pi'_\theta)} \nonumber\\
  & \text{~~~~} P(\Pi|obs_{0:t-1})
\end{align}
Ignoring the denominator (being constant),
\begin{align}
  \pi_{\theta_t}^* &= \operatorname*{argmax}_{\pi_\theta \in \Pi, \Pi \in \mathcal{R}} P( \mathbf{s}_{g_t} | \pi_\theta, obs_{0:t-1}, \Pi) P(\pi_\theta) P(\Pi|obs_{0:t-1}) \label{eq_map_given_task}
\end{align}

Equation \ref{eq_map_given_task} gives the MAP estimation of the next elementary policy from the repertoires to be applied on the robot to achieve the current sub-goal $\mathbf{s}_{g_t}$ in one-step. 

Now, $P(\pi_\theta)$ is the prior belief over the elementary policies. One option is to set it equal for all the policies in the repertoires. However, setting equal (positive) probability for the ones that have transition $\Delta \mathbf{s}_\theta$ in the neighborhood of the desired transition ($\mathbf{s}_{g_t} - \mathbf{s}_{t}$) and setting zero for the others will improve the optimization time by eliminating the need to evaluate all the policies in the repertoires. One thing to be noted here that taking a very small neighborhood might degrade the performance of the algorithm.

$P( \mathbf{s}_{g_t} | \pi_\theta, obs_{0:t-1}, \Pi)$ is the Gaussian likelihood of the transition to $\mathbf{s}_{g_t}$ given the repertoire $\Pi$ (i.e GP with prior mean function from $\Pi$), observations $obs_{0:t-1}$ and the elementary policy $\pi_\theta$. This can be computed using the mean and variance prediction of the GP transformation model learned using $obs_{0:t-1}$ with the repertoire $\Pi$ as mean-function. Here, the input to the GP is $\Delta \mathbf{s}_\theta$, which is the task-space transition corresponding to $\pi_\theta$ in the repertoire $\Pi$. To be more precise, if $\mu(\Delta \mathbf{s}_\theta)$ and $\Sigma(\Delta \mathbf{s}_\theta)$ are the mean and the diagonal covariance predicted by the GP model for an $n$ dimensional task-space, then $P( \mathbf{s}_{g_t} | \pi_\theta, obs_{0:t-1}, \Pi)$ is computed as follows:
\begin{align}
  &P( \mathbf{s}_{g_t} | \pi_\theta, obs_{0:t-1}, \Pi) = \frac{1}{(2\pi)^{n/2}|\Sigma(\Delta \mathbf{s}_\theta)|^{1/2}} \nonumber\\ 
  &\text{~~}\exp \Big(-\frac{1}{2}(s_g - \mu(\Delta \mathbf{s}_\theta))^{T}\Sigma^{-1} (s_g - \mu(\Delta \mathbf{s}_\theta)) \Big)
\end{align}

$P(\Pi | obs_{0:t-1})$ represents the likelihood of a repertoire being able to represent the reality for the robot. To compute this quantity, first we define a closeness score $\psi_{\Pi}$ that represents how close the mappings of the repertoire $\Pi$ are compared to real world observations.
\begin{align}
  \psi_{\Pi} = e ^ { -k||\Delta\mathbf{s}_\theta - \Delta\mathbf{s}_{\theta, real}||^2},  k > 0 \label{eq_closeness}
\end{align} 

where, $\Delta\mathbf{s}_{\theta, real}$ is the observed transition on the robot after applying a policy $\pi_\theta$ taken from the repertoire $\Pi$ and $\Delta\mathbf{s}_\theta$ is the corresponding transition stored in the repertoire. Now, for a real robot, $\Delta\mathbf{s}_{\theta, real}$ can be stochastic for a given policy. Moreover, for different policies from the same repertoire $\psi_{\Pi}$ can be different. This makes $\psi_{\Pi}$ stochastic in nature. Thus, the overall score of the repertoire can be defined as the expectation of $\psi_{\Pi}$. However, to compute a good estimate of the true expectation, it will require several observations from all the repertoires, which will make the adaptation process slow. On the other hand, imperfect estimation of the expectation of $\psi_{\Pi}$ with small number of observations from the repertoires might make the algorithm greedy towards any repertoire that, by chance, has given higher $\psi_{\Pi}$ for the selected policies.
To have a balance between exploration and exploitation of these repertoires, we borrow the concept of \emph{Upper Confidence Bound} (UCB) from the multi-armed bandits problem formulation~\cite{sutton1998reinforcement}. Thus, instead of estimating the expectation by taking the mean of the scores, we compute the UCB as follows:
\begin{align}
  \psi_{ucb\Pi } = \frac{\sum \psi_{\Pi}}{N_{\Pi}} + m\sqrt{\frac{\ln(n)}{N_{\Pi}}} \label{eq:ucb}
\end{align}
Where, $n$ is the total number policies executed on the robot so far, $N_{\Pi}$ is the number of times the policies were used from the repertoire $\Pi$ and $m$ is a positive constant. Note that since we use UCB1~\cite{auer2002finite}, theoretically $m=\sqrt2$. However, we left $m$ as a tunable hyperparameter of \algo{} for specifying the amount of exploration. Normalizing these UCB scores will give higher probability value to those repertoires which have higher ``mean score'' and to those repertoires that are not tried enough compared to others on the real robot:
\begin{align}
  P(\Pi | obs_{0:t-1}) = \frac{\psi_{ucb\Pi}}{\sum_{\Pi' \in \mathcal{R}} \psi_{ucb\Pi' }}
\end{align}

Combining everything, at every time-step, optimizing the equation \ref{eq_map_given_task} gives the optimal policy to be used for the current sub-task. Since our policy space is discrete and equation \ref{eq_map_given_task} is fast to evaluate (it does not involve the simulator), we can simply evaluate all the elementary policies $\pi$ from all the repertoires to find out the optimal according to equation \ref{eq_map_given_task} (see Algorithm~\ref{algo:planning})

\section{Experimental Results}
We evaluate \algo{} on two simulated tasks: (1) object pushing task with a robotic arm and (2) goal reaching task with a damaged hexapod. Additionally, we demonstrate the effectiveness of \algo{} on a damaged 6-legged robot (hexapod) which has to reach the target position in an arena as quickly as possible by avoiding obstacles in the path. We evaluated the following baselines for both the tasks and compared the results to \algo{} with 40 replicates:

\begin{itemize}
  \item CP-L (Close Prior with Learning): Using \algo{} with only one repertoire that is very close to the reality. For example, in the object pushing task, if the test object is a cube, then the repertoire used in this case can be of a cuboid or a slightly larger or slightly smaller cube. For the hexapod task, the floor friction used for the repertoire can be close to the floor friction during test time. Another option is to use a repertoire for 2 blocked legs and at the test time hexapod has only one of those legs blocked. Since CP-L is basically \algo{} with the closest prior to the real situation, it is the best \algo{} can be expected to perform with multiple repertoires. So, we want \algo{} to perform as close as possible to the CP-L baseline in the experiments.

  \item SP-L (Single Prior with Learning): Using a randomly chosen prior repertoire for learning. Here, the chosen repertoire need not necessarily be closer to the reality. In the Hexapod task, this baseline is exactly RTE as we used the author provided implementation of RTE. However, for the object pushing task, we used \algo{} with a randomly chosen repertoire. Thus, in this task SP-L is close to RTE in the sense that (1) both use a single repertoire, (2) both transform this repertoire according to the observed data using a probabilistic model and (3) incorporate the probability estimates to decide the next controller to be applied on the robot. 

  \item SP-NL (Single Prior, No Learning of model) In this baseline, we use \algo{} with one randomly chosen repertoire for adaptation and ablate the learning of transformation model. Without updating the transformation model using observation from test time, it assumes that the chosen repertoire perfectly matches with test object/damage (i.e., the reality).  

  \item APROL-NL (APROL with No Learning): In this baseline, we use \algo{} with several prior repertoires. However, we ablate the learning of transformation model. Thus, it assumes that one of the repertoire will perfectly match to the test object/damage. However, this assumption is not true since we do not include the repertoire that matches with the test object/damage.  
\end{itemize} 

All the simulated experiments were implemented in python (see APPENDIX). For both the tasks, we used the pybullet physics simulation library~\cite{coumans2013bullet}. For comparison with RTE~\cite{chatzilygeroudis2018reset} in the hexapod task we used the author provided code {\url{https://github.com/resibots/chatzilygeroudis_2018_rte}}. For the GP model, we used gpy library~\cite{gpy2014}. A video of the experiments is available here : \url{http://tiny.cc/aprol_video}.

\subsection{Object Pushing with a Robotic Arm}

\begin{figure}
  \centering
  \includegraphics[width=0.7\linewidth]{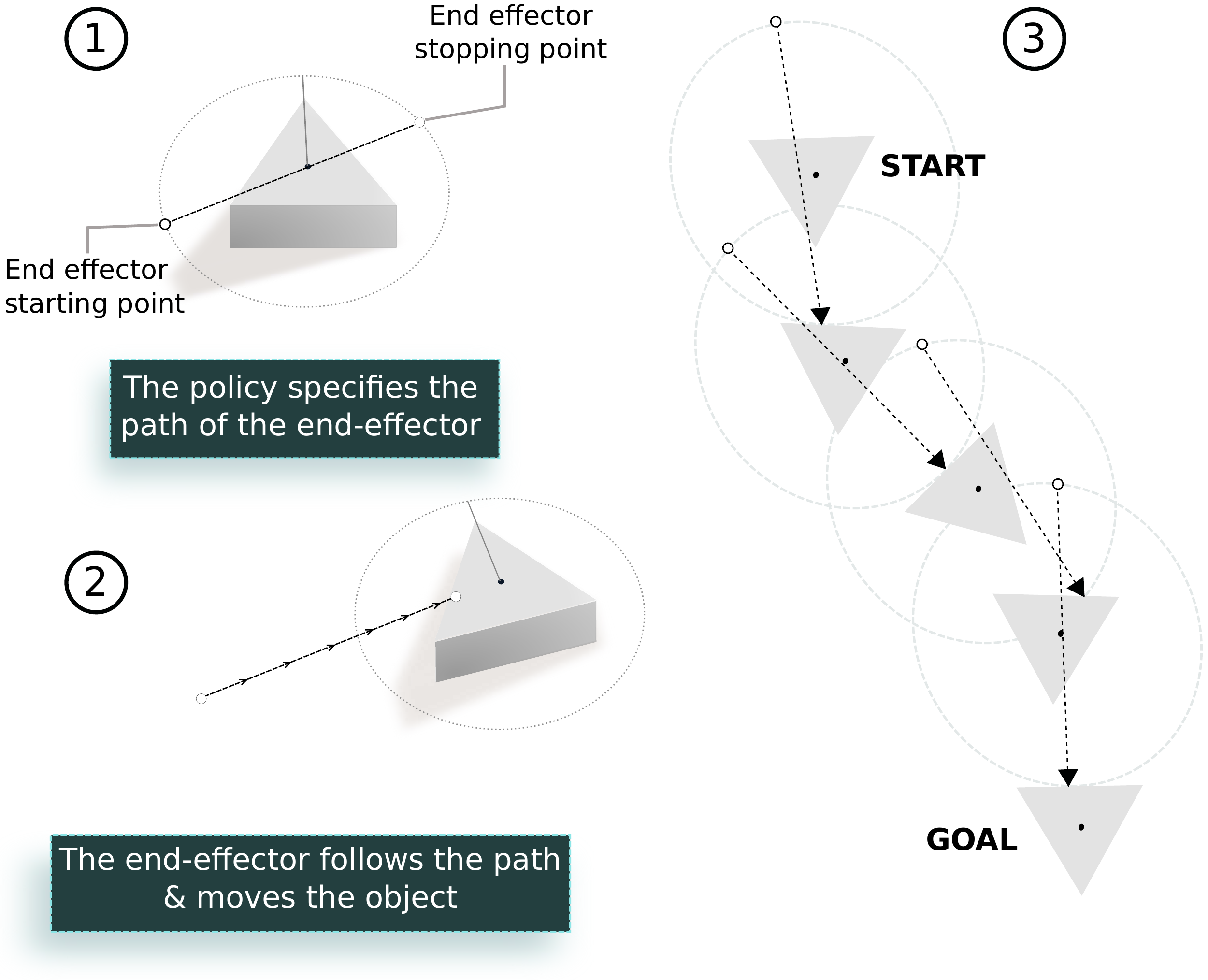}
  \caption{\small\label{fig_4_pushing_policy} The elementary policy for the object pushing task: (1) A 2D vector specifies the start and end position of the end-effector on the surface around the object. The first element of the vector specifies the angular position of the starting point on a circle around the object relative to its current orientation. Similarly, the second element specifies the final end effector position on the same circle. (2) During execution, the end effector follows the straight line connected by these two points using inverse kinematics of the arm. (3) The object can be moved to the goal position by sequencing multiple such policies.}   
\end{figure}

\begin{figure}
  \centering
  \includegraphics[width=0.7\linewidth]{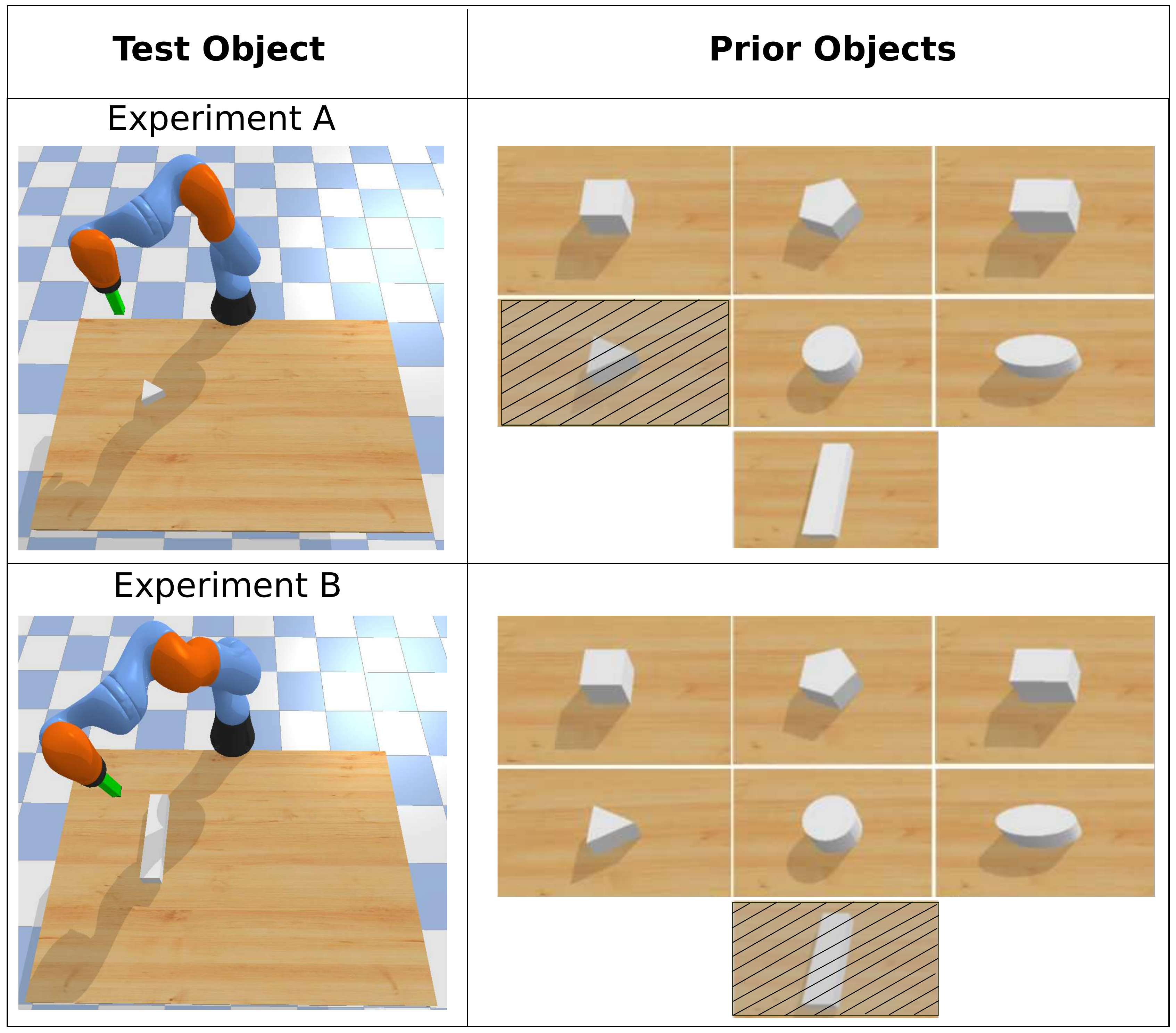}
  \caption{\small\label{fig_5_object_priors} Priors used in object pushing task with \algo{}. Note that in test time with \algo{}, we do not use the exact repertoire (crossed objects in the image) that matches with the test object.}   
\end{figure}

\begin{figure*}[h!]
  \centering
  \includegraphics[width=0.6\linewidth]{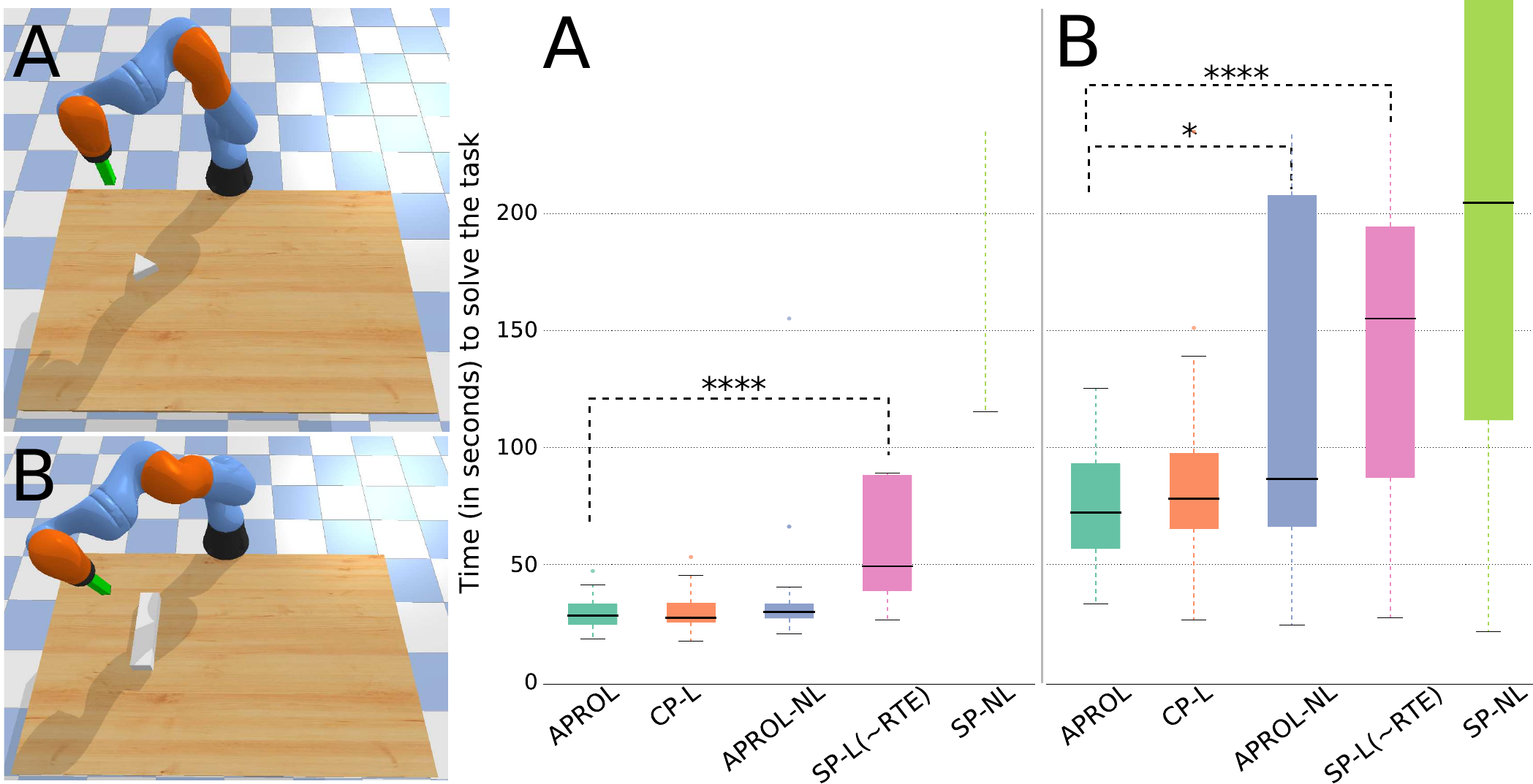}
  \caption{\small\label{fig_6_object_pusher_plots_and_image} Comparison of \algo{} with various single prior and multi-prior variants on an object pushing task with a robotic arm. The goal is to push the objects of various shapes and sizes to different goal positions as quickly as possible. Here, CP: very Close Prior to the reality, SP: Single random Prior, L: with learning the transformation model using GP, NL: Not learning the transformation model with real observations and simply using the expected transitions stored in the repertoire. Asterisks in the plot represent statistical significance (p-value) using Wilcoxon-Mann-Whitney test. For example, 4 asterisks represent $p < 0.0001$, 3 asterisks represent $0.0001 \leq p < 0.001$, 2 asterisks for $0.001 \leq p < 0.01$ and 1 asterisk for $p<0.05$. Higher number of asterisks signifies higher statistical significance between two box plots. (A) With a triangular shaped test object, \algo{} performs at least as good as using a very close prior to the test object (CP-L) and outperforms all the single repertoire based baselines. (B) With a bar shaped test object, here \algo{} performs better than APROL-NL, which does not learn the transformation models with observed data. Also, \algo{} significantly outperforms single repertoire-based baselines.}  
\end{figure*}

The goal here is to push objects of various shapes and sizes to different goal locations in a minimum number of steps. In this task, we assume that the robot has access to its model (so that it can be controlled in Cartesian space) and to the center and orientation of the object from a vision system (for instance, a QR code on the object). However, the robot does not have any knowledge about the shape and size of the objects. The objective is to adapt to push these objects of unknown shapes and guide them to the target position.

\textbf{Elementary policy:} We encode the elementary policy (open loop controller) of the robot with 2 parameters in $[0,1]$. These two parameters specify a straight line connecting two points around the center of the object taking into account the orientation of the object. For given control parameters, the robot's end effector follows the line specified by the parameters in 2 seconds (see Figure~\ref{fig_4_pushing_policy}).

\textbf{Policy-repertoires:} We pre-generated policy repertoires for 7 different objects (Figure~\ref{fig_5_object_priors}) using the MAP-Elites algorithm in simulation. To evolve these repertoires, we did not assign any performance score for the policies. Since the goal of the task is to reach different positions on the 2D surface, therefore, the task-space is the 2D coordinates on the plane. In the repertoires, every policy corresponds to a unique task-space transition of the objects on this plane. \emph{Note that we exclude the exact repertoire that matches with the test object in all the experiments except for CP (very close prior) variants}. For this task, MAP-Elites evaluated 300,000 policies to generate each of the repertoires. Thanks to the cluster of computers, several repertoires could be evolved in parallel in approximately 5 hours of computation.

\textbf{Execution:} At every replanning-step (2 seconds), the robot uses the A* path planning algorithm to plan a shortest sequence of sub-goals in the task space to reach the goal. Then it attempts to achieve the first sub-goal by picking the optimal elementary policy from one of the repertoires using \algo{}. After every step, the robot updates its transformation models using all the past observations. We did not optimize the hyperparameters of the GP here. This is because GPs get only a few data points to learn the model and for small number of data points hyperparameter optimization in GP often doesn't give good results. We set $\sigma_{se}=0.03$ and $l=0.3$ for the kernel function (Eq.\ref{eq:kernel}). These steps are iteratively performed until the object reaches its final goal.

Fig.~\ref{fig_6_object_pusher_plots_and_image}A and Fig.~\ref{fig_6_object_pusher_plots_and_image}B shows that \algo{} performs as good as that of using a very close prior (CP-L) to the reality, i.e., the test object. This shows that \algo{} is able to learn to adapt the transformation model according to the best prior quickly, which allows it to pick policies that align with the desired transitions on the task-space.

\algo{} outperforms the baseline APROL-NL (task B), which is the ablated version of \algo{} by removing the model learning phase. It shows the importance of the online adaptation of the repertoires in \algo{} using the learned GP models. Again, for both the test objects, \algo{} outperforms the single prior variants with and without learning the transformation models (SP-L, SP-NL). 

\subsection{Goal Reaching Task with a Damaged Hexapod}

\begin{figure*}[h!]
  \centering
  \includegraphics[width=0.6\linewidth]{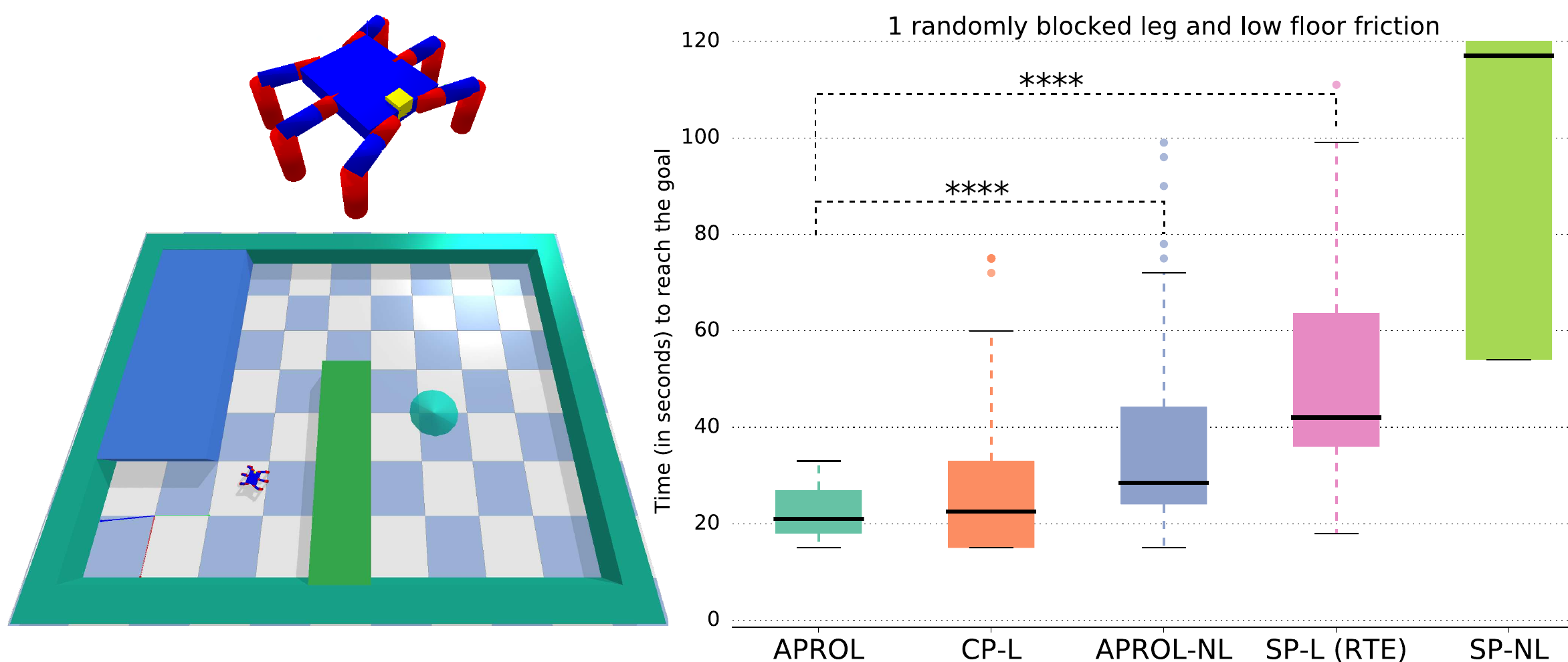}
  \caption{\small\label{fig_7_hexapod_goal_reaching_plot_and_image} Comparison of \algo{} with different single prior and multi-prior variants on simulated hexapod goal reaching task. In this task, the hexapod has to recover from random leg damages and reduced floor friction to reach the goal as quickly as possible. Here, CP: very Close Prior to the reality, SP: Single random Prior, L: with learning the transformation model using GP, NL: not learning the transformation model with real observations and simply using the expected transitions stored in the repertoire. Asterisks represent p-value using Wilcoxon-Mann-Whitney test. For example, 4 asterisks represent $p < 0.0001$, 3 asterisks represent $0.0001 \leq p < 0.001$, 2 asterisks for $0.001 \leq p < 0.01$ and 1 asterisk for $p<0.05$. Higher number of asterisks signifies higher statistical significance between two box plots. In this experiment, \algo{} performs as good as that of using a very close prior to the real situation (CP-L) and outperforms all the single repertoire based baselines.}  
\end{figure*}

\begin{figure}
    \centering
    \includegraphics[width=0.45\textwidth]{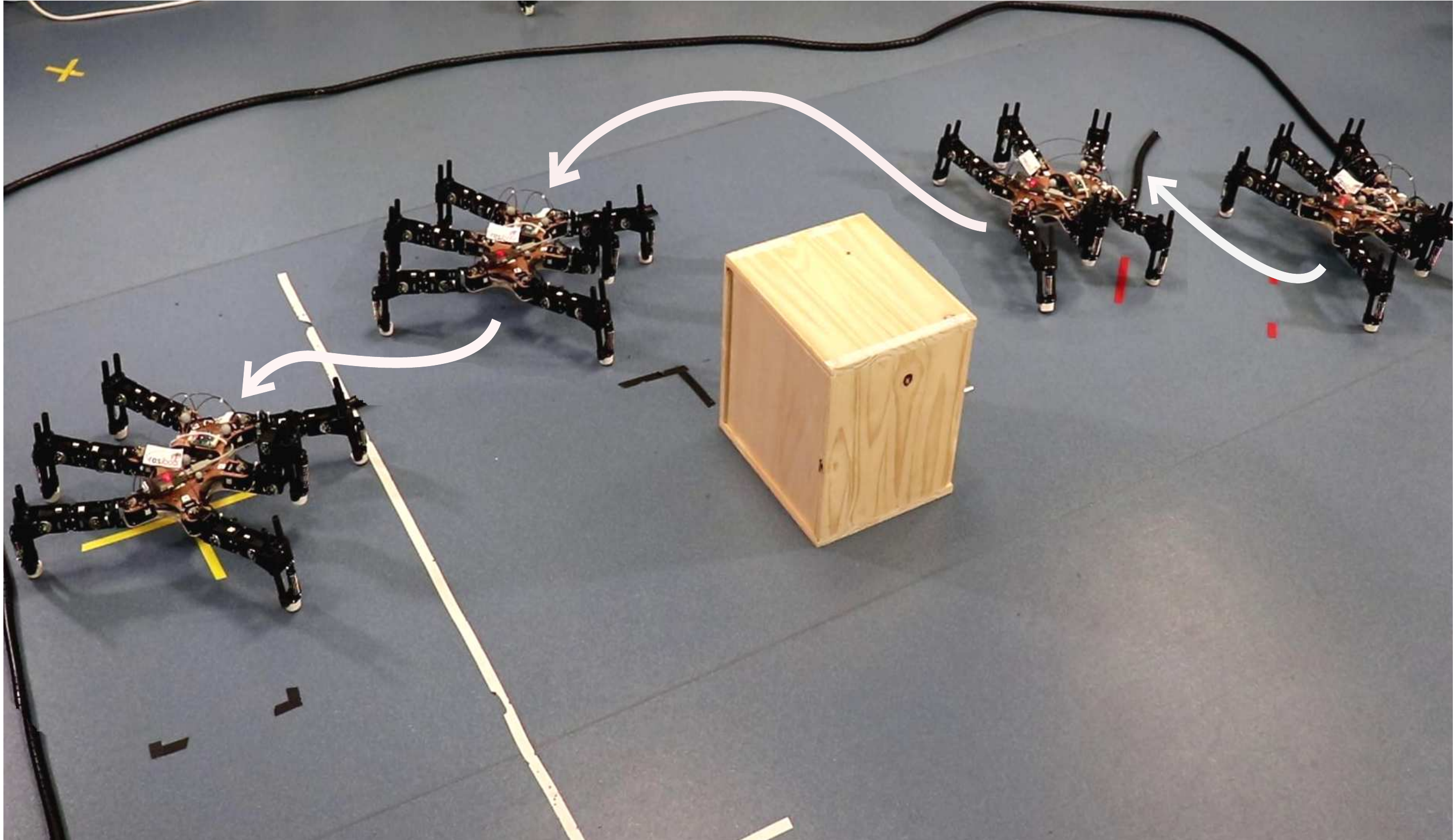}
  \caption{\small\label{fig_8_real_hexa_v3} Goal reaching task with a real hexapod with blocked leg.}  
  \vspace{-1em}
\end{figure}

This task is performed both in simulation and on a real hexapod robot. In this task, the robot might encounter various situations, such as, damage to one or more legs (e.g, blocked joints or a lost leg) and various friction conditions of the floor (e.g., very low, moderate and very high friction co-efficient). The goal here is to reach the specified position in minimum number of control-steps. Here, the robot has the knowledge of its position and orientation (e.g., from SLAM or visual odometry system). Additionally, we assume that the robot has the complete knowledge of the obstacle positions as well as dimensions. However, the robot does not have any knowledge about the damage to its legs and floor friction condition. 

\textbf{Elementary policy:} The robot has 6 identical legs, each of which has 3 degrees of freedom (DOF). The first DOF ($\theta_0$) controls the horizontal movement of the leg and the seconds and the third DOFs ($\theta_1 \text{ and } \theta_2$) control the elevation of the leg. $\theta_2$ is set equal to the negative of $\theta_1$ always to make the final segment of the leg parallel to the body. Therefore, there are $6 \times 2 =12 $ independent joints in the robot. Each of these joints are controlled with 3 parameters: the amplitude, the phase, and the duty-cycle which are used in a periodic function of time to produce joint positions. These $3 \times 12=36$ parameters define the elementary policy for the robot (see~\cite{cully_robots_2015} for more details about this controller).

\textbf{Policy repertoires:} Before deployment, policy repertoires for various situations were generated for the robot using MAP-Elites in simulation. More precisely, we generated repertoires for 3 different friction co-efficients $(0.6, 1.0, 5.0)$ of the floor and various leg damage conditions (single and two leg damages/blocks). Out of $108$ possible combinations, we selected randomly $57$ combinations to generate the repertoires for the hexapod. Each repertoire contains discrete mappings from 36D policy to 2D task-space transitions of the robot on the surface. Note that, in this task, the task space is simply the center of mass position of the robot. Thus, repertoires have transitions of the center of mass of the robot for different policies. For this task, MAP-Elites evaluated 500,000 policies in simulation to generate each of the repertoires. Thanks to the cluster of computers, several repertoires could be evolved in parallel within 12 hours of computation. 

\textbf{Execution: } At every replanning-step (3 seconds), the robot uses the A* planning algorithm to plan a shortest sequence of positions that the robot has to follow to reach the goal by avoiding any obstacle in the path. Then it tries to reach the first sub-goal by selecting and executing the most suitable elementary policy from one of the repertoires using \algo{}. After every execution, the transformation model for the repertoire from which the policy has been selected is updated with all the previously observed data. Similar to the previous experiment, we did not optimize the hyperparameters of the GP here due to small number of data. We set $\sigma_{se}=0.03$ and $l=0.3$ for the kernel function (Eq.\ref{eq:kernel}) in this task. After each execution (3 seconds duration), the robot re-plans the sequence of positions using A*. The process continues until the robot reaches the goal position. 

In the simulated hexapod goal reaching task, we evaluated each of the variants with 40 replicates. Fig.~\ref{fig_7_hexapod_goal_reaching_plot_and_image} shows that \algo{} performs at least as good as that of using a very close prior (i.e., CP-L variant) to the reality (i.e., the exact leg damage and very similar floor friction). Note that with \algo{} we did not include the repertoire that exactly matches with the real situation of the robot. This suggests that \algo{} is able to quickly figure out the most suitable repertoire to use it as prior for learning the transformation model, and from which it accordingly selects the most suitable policies. 

\algo{} outperforms the baseline that uses multiple priors but without any learning of the transformation models (APROL-NL). That is, in APROL-NL, we assume that one of the repertoires will exactly match the reality. However, since this assumption is not true, \algo{} performs better than this baseline. Additionally, \algo{} outperforms the single prior baseline with and without learning the transformation models (SP-L, SP-NL). Note that, SP-L is exactly RTE in this task.

Additionally, we have demonstrated the capability of \algo{} in a real hexapod damage recovery and goal reaching task (Figure \ref{fig_8_real_hexa_v3}) with total 8 replicates. We show that, with \algo{}, the damaged robot learns to select the compensatory policies to reach the goal by avoiding the obstacle in the path. Here the robot reaches the goal positions by avoiding obstacle (a wooden box) with $100\%$ success (within maximum 30 steps). Compared to the intact robot (with single repertoire generated for the intact robot itself), the damaged robot was able to recover its capability upto $88\%$ in terms of the time it took to reach the final goal. One thing to be noted here that this was a challenging task not only due to the damage, but also due to the reality gap between simulation and the reality. As a result, in this task even the exact prior repertoire gives very high mismatch in the behavior when applied in the simulation and on the real robot. A video of all the experiments is available here: \url{http://tiny.cc/aprol_video}.  

\section{Discussion and Conclusion}

Prior knowledge is key to rapid adaptation, be it in repertoire-based learning, meta-learning, or model-based policy search. However, the effectiveness of the prior knowledge is highly dependent upon how relevant it is to the current scenario or situation that the robot is facing during deployment. In fact, a wrongly chosen prior might hinder or prolong the learning process instead. For example, a policy repertoire that is generated for a robot to walk on a flat surface is not a good prior for a robot that has to walk on stairs \cite{pautrat2018bayesian}. Unlike \cite{chatzilygeroudis2018reset}, in this work we relax the assumption that a single repertoire-based prior will be able to capture all the situations. Instead, we allow the robot to choose the best prior among many repertoire-based priors to achieve faster adaptation. It is to be noted that for episodic learning, \cite{pautrat2018bayesian} also reached a similar conclusion.

We believe APROL can find its application in many different fronts, such as fault tolerance or damage recovery in robotics, adaptation to sudden changes in environmental conditions, transferring controllers learned in simulation to the real robot etc. For example, in case of robots deployed in places (e.g., space, deep sea, radio-active zones) where a human has to control them remotely with high level commands (e.g., move in different directions, grab or push object, etc.), the built-in controllers may not give the desired effect if any fault occurs in their joints. In such situations, using APROL, a robot can learn to use alternative controllers to accomplish the command. Again, for complex robots, a policy learned in simulation often gives slightly different outcomes on the real robot. In such situations also, APROL can learn to pick alternative policies to have the desired outcome. 

Compared to model-based reinforcement learning \cite{deisenroth_gaussian_2015,chatzilygeroudis2017black}, \algo{} is faster as it does not perform optimization on the policy parameter space. Instead, \algo{} learns to ``select'' the most suitable elementary policy from the given repertoires according to the current situation and the goal. However, \algo{} has to evaluate the outcomes of all the policies stored in the repertoires using the Gaussian processes model, which has cubic time complexity. Therefore, with more and more data points, optimization of the policy between two replanning steps of the robot becomes slower and slower. To mitigate this problem, we can incorporate several strategies. One option is that instead of learning the GPs from all the past observations, they can be learned from $M$ recent observations, where $M$ can be a hyperparameter based on the desired optimization speed that we want to achieve. Learning from $M$ recent observations will also allow the robot to continue its mission if multiple changes in the situation occur over the time (e.g., plane surface, then very rough surface and then very slippery surface). Another option is to use sparse GPs~\cite{quinonero2005unifying} or local GPs~\cite{park2017patchwork} or neural networks with uncertainty~\cite{gal2015dropout}. Another important thing to be noted here that in this work we assumed that the task-space is much smaller than the full state-space of the system. This assumption is true in most of the cases in robotics. We do not expect the algorithm to scale for problems with very high dimensional task-space. This is because the number of observations required to learn the models will grow exponentially with the task-space dimension causing adaptation time longer. However, we can expect \algo{} to scale well to problems with very high dimensional state-space as long as their task-space is smaller (e.g less than 10 dimensional).

There are several hyperparameters associated with \algo{} and the most of them are linked to the MAP-Elites algorithm and GP models. One of the most important parameter that is associated with MAP-Elites is the amount of discretization of the task-space. Since we used the CVT variant of MAP-Elites~\cite{vassiliades2017using}, we can specify the number of cells we want to generate after the discretization of the space. Now, a coarse discretization can be detrimental for \algo{} since there will be less number of elementary policies in the repertoire and hence there will be less diversity. On the other hand, a very fine discretization will produce lots of cells in the space and thus there will more diverse policies in the repertoire which will help \algo{} for better adaptation. However, higher number of cells will require higher number total evaluations in Map-Elites. As a result, it will increase the time required to generate the repertoires in simulation. Again, since \algo{} evaluates potentially all the policies from all the repertoires at every replanning step, higher number of policies means higher optimization time. Another important decision in \algo{} is the number of repertoires. If the number of repertoires is small, then they will represent less diverse situations that the robot might face. As a result, if the real situation of the robot is not close to any of the repertoires, then \algo{} might not show any significant improvement over using just a single repertoire. On the other hand, taking a large number of repertoires might prolong the adaptation time. In the situation where the number of repertoires is very high, the exploration parameter $m$ in the equation~\ref{eq:ucb} will play a major role in deciding the adaptation time. A higher value of $m$ will encourage more exploration of the repertoires, thereby prolonging the adaptation time.

One limitation of \algo{} (as well as repertoire-based learning such as RTE) is that instead of learning a transition model in full state-space of the robot, it learns a transformation model of the task-space assuming that task-space transition is independent of the current state of the robot. This assumption is not always true (e.g, for the end effector of a robotic arm). However, in both our experiments, this assumption holds. This is because, for the hexapod, we reset the joints between each replanning step of the robot. That makes the effective full state of the robot equal to position and orientation only, which are independent of each other for the hexapod since our repertoires consider only the ``change in position'' from the current position. Similarly, for the object pushing task, we consider both position and orientation of the object, which is the same as that of the full state of the object. Thus, \algo{} is more suitable for mostly robot locomotion tasks as well as tasks where, to some extent, the above mentioned assumption holds. 

In spite of using a finite set of policies for adaptation or learning, so far, repertoire-based approaches have been able to show many promising results in real robotic systems. Thanks to evolutionary algorithms such as MAP-Elites, the key element of such promising results is the diversity of the policies stored in the repertoires. As it happens in nature, due to this diversity, many such policies can still ``survive'' (i.e., work on the robot) even if any catastrophic event (such as joint failure) happens during the mission. We believe, repertoire-based adaptation algorithms such as \algo{} will open new frontiers in the direction of rapid adaptation for robotic systems in the real and uncertain world.



\section*{Appendix}
\label{sec:appendix}
Code for this work can be found in the following github repository: \url{https://github.com/resibots/kaushik_2019_aprol} 
\bibliographystyle{IEEEtran}
\bibliography{mybib}

\end{document}